\documentclass[math]{MSword}

\usepackage[english]{babel}
\usepackage{csquotes}
\usepackage[pagewise, mathlines]{lineno}

\newcommand{\institution}[1]{{\footnotesize\color{black}#1}} 

\usepackage{titling} 

\pretitle{\vspace{-20mm}}%

\posttitle{\par\vskip1pt \vspace{0mm}} 

\preauthor{} 

\postauthor{\vspace{-14mm}}

\title{\Large \textbf{Simmering: sufficient is better than optimal for training neural
networks}}
\author{Irina Babayan,\textsuperscript{1$\dagger$}
  Hazhir Aliahmadi\textsuperscript{1$\dagger$},
  and
  Greg van Anders\textsuperscript{1*} 
	\newline\newline 
    \institution{\textsuperscript{1}\textit{
    Department of Physics, Engineering Physics,
  and Astronomy, Queen's University, Kingston ON, K7L 3N6, Canada}}\\
  \institution{\textsuperscript{$\dagger$}\textit{These authors contributed equally.}}
  \institution{\textsuperscript{*} \textit{gva@queensu.ca}}
}
\date{}
\usepackage[backend=biber,bibencoding=utf8,style=nature,sorting=none,natbib=true, defernumbers=true]{biblatex}
\DeclareBibliographyCategory{methods}
\addtocategory{methods}{PINNs}
\addtocategory{methods}{nealInference1993}
\addtocategory{methods}{wrightBayesianNN1999}
\addtocategory{methods}{GlennMartyna1996}
\addtocategory{methods}{Tuckerman1992}
\DeclareBibliographyCategory{si}
\addtocategory{si}{deng2012_mnist}
\addtocategory{si}{whitesonHIGGS2014}
\addtocategory{si}{iris_53}
\addtocategory{si}{lecunGradient1998}
\addtocategory{si}{quinlanAutoMPG1993}
\let\cite=\supercite

\begin{document}

\maketitle

\begin{refsegment}

\noindent{\textbf{%
  The broad range of neural network training techniques that invoke optimization
  but rely on ad hoc modification for validity \cite{precheltEarlyStopping1998, Bagging,  freundBoosting1996, Dropout,matsuokaNoiseInjection1992,hoWeightNoiseInjection2009,tibshiraniRegression1996,polikarEnsembleLearning2012,hastieStatisticalLearning2009} suggests that
  optimization-based training is misguided. Shortcomings of optimization-based
  training are brought to particularly strong relief by the problem of
  overfitting, where naive optimization produces spurious outcomes.\cite{hawkins2004_problem_overfitting,bilbao2017_overfitting,yingOverfitting2019}
  The broad success of neural networks for modelling physical
  processes \cite{Kutz2017_fm,pearsonCNN2019,shiThermo2021,marcatoFluidDynamics2022,tf} has prompted advances that are based on inverting the
  direction of investigation and treating
  neural networks as if they were physical systems in their own right.\cite{sohl2015deep,rotskoffTrainability2022,huDiffusion2018,mei2018mean}
  These successes raise the question of whether broader, physical perspectives
  could motivate the construction of improved training algorithms.
  Here, we introduce simmering, a physics-based method that trains neural
  networks to generate weights and biases that are merely ``good enough'', but
  which, paradoxically, outperforms leading optimization-based approaches.
  Using classification and regression examples we show that simmering corrects
  neural networks that are overfit by Adam \cite{Adam}, and show that simmering
  avoids overfitting if deployed from the outset.
  Our results question optimization as a paradigm for neural network
  training, and leverage information-geometric arguments to point to the
  existence of classes of sufficient-training algorithms that do not take
  optimization as their starting point.
}}

\section*{Main}
Although neural networks' universal estimation capability
\cite{hornikMultilayerFFNN1989,cybenkoApproximation1989,luExpressive2017,guliyevApprox2018}
allows them to represent many complex data
relationships,\cite{DeepLearning} that capability makes training
generalizable networks challenging. The over-parameterization that supports the
universal capabilities of neural networks nonetheless gives key advantages over
other estimators in settings where complex data produce a training loss
landscape that is non-convex and has many local
minima.\cite{livniComputationalEff2014, weiRegularization2019} Yet noise in
training data can misdirect the parameter estimation process towards an
overspecified representation that accurately respects idiosyncrasies in training
data, but that severely limits
generalizability.\cite{mehrabiFairness2022,frenayClassification2014,
torralbaUnbiased2011} This accuracy-generalizability discord is
exacerbated by optimization-based training methods, which are overly effective
at exploiting universal estimation capacity to achieve minimized-loss
representations of training data idiosyncracies.

The danger of combining the excessive expressiveness of a neural network
and discrepant data with optimization is brought to particular relief by
overfitting.\cite{ruderOverviewGD2017,ghojoghTheoryOverfitting2023}
Overfit neural networks are inevitable when an over-parameterized architecture
is combined with an efficient optimization algorithm
\cite{livniComputationalEff2014} (e.g., Adam
\cite{Adam}). Efficient optimization
yields high-complexity networks that generalize poorly because
optimization-based training cannot distinguish between the ``ground truth''and
the noise in the data during training. Attempts to mitigate overfitting, e.g.,
early stopping,\cite{precheltEarlyStopping1998} bagging,
\cite{Bagging} boosting,\cite{freundBoosting1996}
dropout,\cite{Dropout} all account for data uncertainty by
incorporating deviations from empirical error minimization into training.
However, the effectiveness of most overfitting mitigation techniques relies on
the data distribution satisfying specific assumptions,
\cite{ghojoghTheoryOverfitting2023} and is thus problem dependent.
Nonetheless, the success of avoiding overfitting via increased training loss
\cite{ghojoghTheoryOverfitting2023} suggests that more generalizable
representations of ground truth are near-optimal rather than
optimal.\cite{ruderOverviewGD2017} Thus, training paradigms that
are founded on an alternate premise, e.g., sufficiency rather than optimality,
could produce non-overfit, generalizable estimators while still benefiting from
the expressive capacity of neural networks.

Here, we demonstrate that simmering, an example of a sufficient-training
algorithm, can improve on optimization-based training. Using examples of
regression and classification problems learned via feedforward neural networks,
we deploy simmering to ``retrofit'', or reduce overfitting, in networks that are
overfit via conventional implementations of
Adam.\cite{Adam} Our approach leverages
Nos\'e-Hoover chain thermostats from molecular dynamics \cite{frenkelsmit} to
treat network weights and biases as ``particles'' imbued with auxiliary,
finite-temperature dynamics and ``forces'' generated by
backpropagation.\cite{BackPropagation} The
finite-temperature dynamics act as a minimally-biased model of the data noise
that systematically prevents the network parameters from reaching optimal
configurations. We also deploy simmering from the outset, rather than first
optimizing and then retrofitting, to train neural networks and show that, in
addition to yielding generalizable neural networks, simmering also yields
quantifiable prediction uncertainty estimates in regression and classification
problems.

Our retrofitting results indicate that simmering is a viable approach to reduce
the overfitting that is inherent in optimization-based training. To understand
why simmering works, we use information geometry arguments
\cite{ParetoLaplace} to show that simmering is but one of a
family of sufficient-training algorithms that improve on optimization-based
training by leveraging mathematical properties of filters in a way that exploits
generic features of loss function landscapes. Our implementation of simmering, a filter-based neural network
training method, is available open source at Ref.\
\cite{SimmeringCode}. Within the general class of
sufficient learning algorithms, information theoretic arguments indicate that
simmering is one of a family of filter-based algorithms that make
minimally-biased assumptions about the form of deviation from ground truth
present in the training data.
This opens the door to statistical-physics
based sufficient-training approaches, e.g., by leveraging other molecular dynamics algorithms.

\section*{Sufficient Training by Simmering}
Existing, optimization-based training algorithms that work to mitigate
overfitting are engineered to avoid optimizing the empirical error because optimized sets
of weights and biases do not reproduce ground truth in generic problems. The
fact that generalizable representations of ground truth do not optimize the empirical error 
suggests the need to systematically explore non-optimal configurations. Exploring non-optimal configurations in generic problems where 
it is not known a priori how training data depart from ground
truth motivates generating minimally biased deviations from optimality.
Information theory suggests employing a generating function that is the Pareto-Laplace
transform \cite{ParetoLaplace} of the training loss
\begin{equation}
    Z(\beta, \mathcal{D}) = \int d^N x e^{-\beta L(x, \mathcal{D})}
  \label{eq:PLxfm}
\end{equation}
where x is the set of neural network parameters (weights and biases), $\vec{x}=(\vec{w},\vec{b})$, $N$ is the total number of neural network parameters, $L(x, \mathcal{D})$ is the loss function evaluated over the training data $\mathcal{D}$, and $\beta$ is the Laplace transform variable. $Z$ is a generating function for sufficiently trained
networks, and generates networks that minimize training loss in the limit $\beta\to\infty$.

We use Eq.\ \eqref{eq:PLxfm} to generate sufficiently trained networks
algorithmically by identifying $Z(\beta, \mathcal{D})$ as a partition function in statistical
mechanics.\cite{ParetoLaplace} The training algorithm (see Methods) treats
$Z(\beta, \mathcal{D})$ as the thermal, configuration space integral of a system of classical
``particles'' representing the weights and biases of the network, with each particle's
1D motion driven by an interaction potential energy determined by the training loss. In
this representation we take $T=1/\beta$ as the system temperature. Without loss
of generality, we lift this configuration space to a phase space by augmenting
each weight and bias with an auxiliary, canonically conjugate momentum and a
canonical, non-relativistic kinetic energy. These momenta and kinetic energy
impart the system with auxiliary dynamics.

This dynamical approach entails gradient forces on weights and biases that can be
computed via backpropagation. We thermalize the dynamics to
reproduce the distribution Eq.\ \eqref{eq:PLxfm}. We operationalize the
thermalization via numerical integration of the equations of motion of the neural network coupled to a
Nos\'e-Hoover chain thermostat, which we implement (see Methods) by
symplectic integration.\cite{ReversibleNHT,Yamamoto2006} 
For portability, and to facilitate use for problems beyond those we study in detail
below, we implement the algorithm in Python and model the neural network in TensorFlow, leveraging TensorFlow's autodifferentiation to compute the gradient forces that drive training dynamics. An open-source implementation of our
approach is available at Ref.\ \cite{SimmeringCode}.

Eq.\ \eqref{eq:PLxfm} serves as a generating function for sufficiently trained
networks for finite $\beta=1/T$. The key to obtaining sufficient training in the dynamical approach we use here is to 
maintain the auxiliary dynamics at a small but finite temperature, or
``simmer'', so that the network systematically explores near-optimal
configurations of weights and biases. 
These near-optimal configurations need not
individually improve the network accuracy
for test data because they belong to ensembles
of networks, and various methods \cite{polikarEnsembleLearning2012,dietterichEnsemble2000} can be applied to the ensembles to extract more generalizable representations.
We give general arguments in Methods that
this supremacy of sufficient training over
optimal training is a generic feature of
the family of methods we introduce here.

\begin{figure}
    \centering
    \includegraphics[width=120mm]{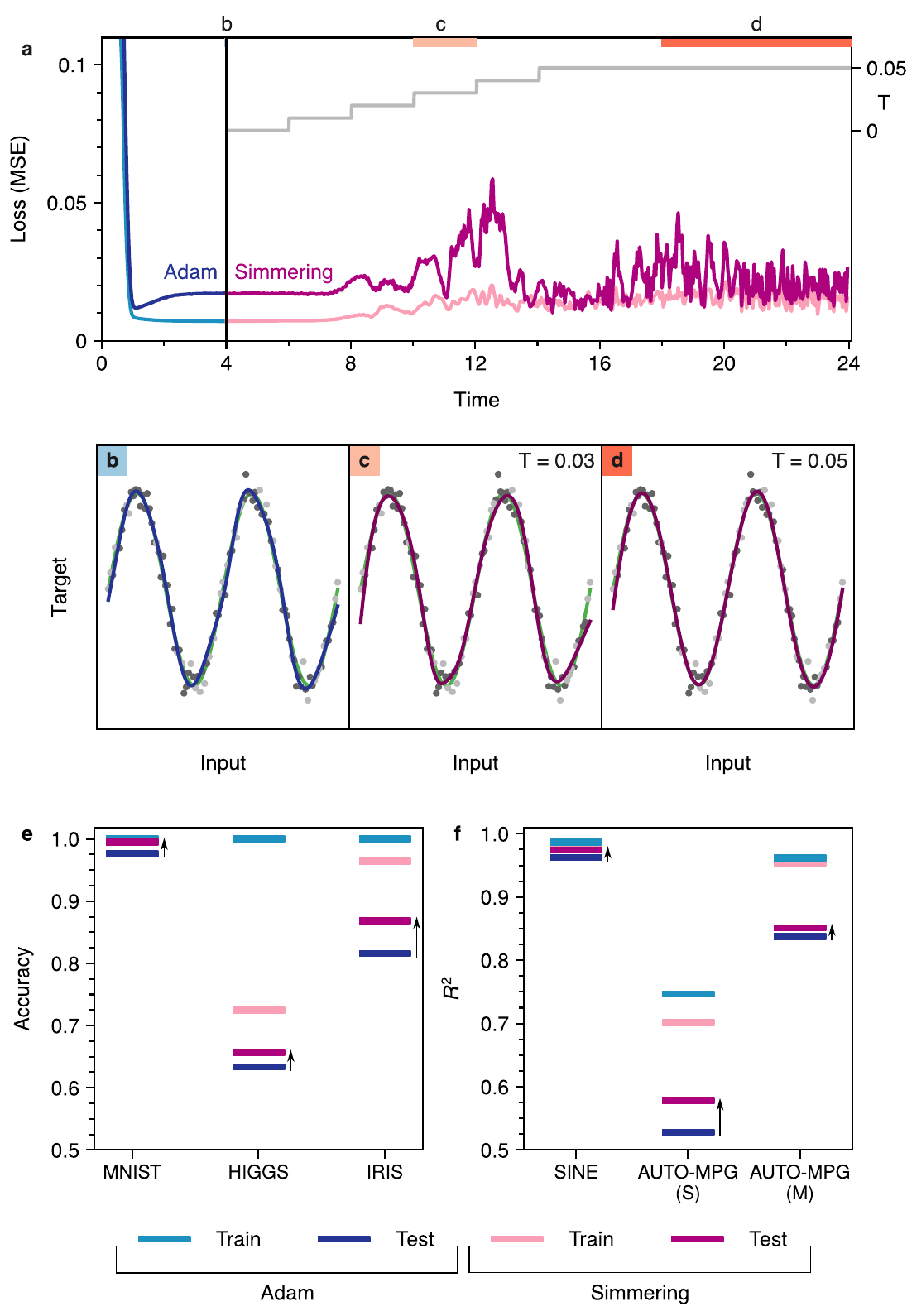}
    \caption{\textbf{Sufficient training based retrofitting reduces overfitting
      in optimized networks.} Optimization-based training produces
      discrepancies in performance on training vs. testing data (c.f.\ light blue and dark
      blue MSE curves, panel a) that manifest in discrepancies between model
      fits and underlying relationships (c.f.\ dark blue and green curves,
      respectively, in panel b). We apply simmering to retrofit the overfit
      network by gradually increasing temperature (c.f.\ grey lines in panel a), which reduces overfitting
      (panel c) before producing an ensemble of networks that yield model
      predictions that are nearly indistinguishable from the underlying
      data distribution (c.f.\ dark magenta and green curves, panel d). Analogous
      applications of simmering can be employed to retrofit classification problems (panel e) and regression problems
      (panel f). Panel e shows prediction accuracy for image classification (MNIST), event
      classification (HIGGS), and species classification (IRIS).
      Panel f shows fit quality (squared residual, $R^2$) for regression
      problems including the sinusoidal fit shown in detail in panels a-d, as
      well as single- (S) and multivariate regression (M) of automotive mileage
      data (AUTO-MPG). In all cases, simmering reduces the overfitting produced
      by Adam (indicated by black arrows). 
    }
    \label{fig:Retro}
\end{figure}

To facilitate comparison between sufficient- and optimal training methods, we
first deploy simmering to reduce overfitting in networks trained using Adam.
Fig.\ \ref{fig:Retro} gives an example of this ``retrofitting'' procedure in
the case of a standard curve fitting problem. Fig.\ \ref{fig:Retro}b shows a
set of training and testing data that are generated by adding noise to a
sinusoidal signal (green line). With these training data, we train the parameters of a fully connected feedforward network using Adam. Fig.\ \ref{fig:Retro}a shows the evolution of the loss, with a
clear divergence of the training and test loss during the Adam training stage. Fig.\ \ref{fig:Retro}b shows
that the Adam-generated fit discernibly deviates from the true signal.

To correct this deviation, we apply simmering, taking the overfit,
Adam-generated network as the initial condition. We introduce step-wise increases
in temperature (grey line, Fig.\ \ref{fig:Retro}a) from $T=0$ to $T=0.05$ (taking $T$ to be measured in units of loss). Simmering generates ensembles of
sufficiently trained networks at finite $T$ which we then aggregate to construct a
``retrofitted'' representation of the underlying signal. Fig.\
\ref{fig:Retro}c shows that simmering has reduced the discrepancies that
were present between the Adam-produced fit and the original signal. Fig.\
\ref{fig:Retro}d shows that a simmering-generated ensemble of sufficiently
trained networks at $T=0.05$ generates an aggregated fit that is virtually indistinguishable
from the original sinusoidal signal.

We carried out an analogous retrofitting procedure on a set of similar problems.
Fig.\ \ref{fig:Retro}e shows retrofitting results for classification problems, where, in all tested
cases, applying simmering to retrofit overfit networks results in improved
classification accuracy on test data. Fig.\ \ref{fig:Retro}f shows results for the
application of simmering to retrofit regression problems where simmering reduces the
residual of the fit for test data compared with overfit, Adam-produced
networks.

\section*{Ab Initio Sufficient Training}

\begin{figure}[ht]
    \centering
    \includegraphics[width=170mm]{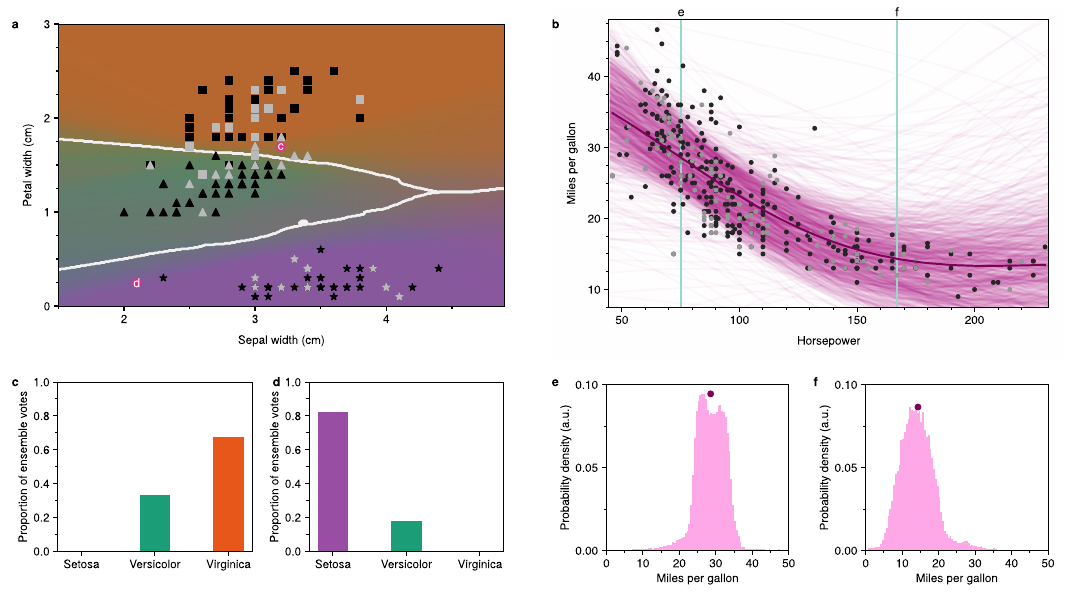}
    \caption{
        \textbf{Ab initio sufficient training avoids overfitting and yields
    prediction uncertainty distributions.} Ensembles of models sampled at finite
    temperature yield smooth decision boundaries (white lines in panel a) and
    average predictions (dark magenta curve in panel b) that are not skewed by
    noisy training data (indicated by black markers in panels a and b). The
    background in panel a is shaded using a weighted average of the ensemble
    votes for each point in the feature space, showing regions of confident
    ensemble prediction (regions of bright orange, teal, or purple in panel a) vs.\
    uncertain prediction (intermediate coloured regions in panel a).
    Analogously, panel b shows the density of predicted curves (transparent
    magenta curves in panel b) around the ensemble average (dark magenta curve
    in panel b). For classification problems, panels c and d show the ensemble's decision-making confidence at different points in the
    data feature space via the proportion of ensemble votes for each
    class (c.f.\ panels c and d correspond to points labelled c and d on panel a). For
    regression problems, we can compare the distributions of sampled predictions with the ensemble average at different input values (c.f.\ pink
    solution distribution and dark magenta point on panels e and f,
    sampled at two different inputs indicated in panel b) and assess how the
    data noise distribution affects predictions throughout the feature space. Ab initio
    sufficient training produces correspondingly sufficiently
    descriptive predictions alongside insight into the
    ensemble prediction
    process that is inaccessible with a single, optimized model. 
    }
    \label{fig:AbInitioSimmering}
\end{figure}

Fig.\ \ref{fig:Retro} demonstrated several applications of simmering to retrofit
networks that are susceptible to overfitting by conventional, optimization-based
training. These results raise the question of whether optimization-based
training is necessary, or whether ``ab initio'' implementations of sufficient
training could avoid overfitting without any need for an optimized initial condition. 

Fig.\ \ref{fig:AbInitioSimmering} shows results from sufficiently trained neural networks in
which simmering was deployed from the outset, without the need for optimization.
Fig.\ \ref{fig:AbInitioSimmering}a shows results for classification and Fig.\ \ref{fig:AbInitioSimmering}b shows results
for regression. It is important to note that
because simmering yields an ensemble of networks, like other ensemble learning approaches, it can be used to generate
prediction uncertainty estimates that mitigate the artificial
precision that arises from singular, optimization-generated solutions. These
uncertainty estimates are shown in Fig.\ \ref{fig:AbInitioSimmering}c-f. A key advantage of simmering is 
that, by exploring the maximum entropy distribution of weights and biases, it makes minimally biased assumptions about the training data error distribution and thus yields a minimally biased prediction uncertainty estimate.

\section*{Generalized Sufficient Training}
The simmering method we presented above demonstrates that sufficient training
consistently generates more generalizable networks than optimization-based
training in all of the cases we tested. To understand why, it is useful to
characterize how neural network architecture and data noise affect the geometric
structure of the loss landscape, and how optimization and simmering traverse
this geometric structure to train generalizable neural networks.

In Methods, we give a detailed argument that the neural network
over-parameterization that drives both universal estimation and overfitting
produces families of parameter combinations with near-equivalent or equivalent
training loss. An optimization algorithm can effectively locate any one of the
redundant minimized training-loss parameter combinations. However, the training loss
landscape geometry is defined by data that generically deviate from ground
truth, so the optimized parameters will always lie some distance in parameter
space from those that describe the underlying phenomenon. This fact means that
optimization-based training is doomed to fail (in terms of generalizability) any
time it works (in terms of approaching optimality).

To understand how to construct training approaches that avoid that fate, it is
instructive to situate the loss landscape of a neural network in the framework
of information
geometry.\cite{transtrumGeometry2011,quinnInPCA2019,OriginSimplicity}
From an information geometry perspective, the families of near-equivalent
training loss networks exist along ``sloppy modes'' in the parameter space,
which can be identified by the spectrum of an appropriate Fisher information
metric.\cite{transtrumGeometry2011} Simmering exploits this feature by
parametrically reshaping the geometry of the loss landscape. Simmering does this
by ``lifting'' optimization algorithms to sample near-optimal sloppy
parameter combinations via a Pareto-Laplace transform \cite{ParetoLaplace}
controlled by a temperature parameter $T=1/\beta$. In Methods, we show that $T$
effectively reduces the distance in parameter space between optimal and
near-optimal parameter sets. By reducing these distances, simmering explores
parameters away from the minimal training loss features formed by dataset
idiosyncrasies. Instead, by systematically deviating from minimized loss solutions,
simmering encounters more generalizable data representations. Simmering
implements this parameter sampling approach by connecting the Pareto-Laplace
``filter'' to molecular dynamics\cite{ParetoLaplace} and taking advantage of the
temperature regulation features of Nos\'e-Hoover thermostats,\cite{frenkelsmit}
but other thermostat algorithms could also exploit this effect to implement
sufficient training.

Accuracy--generalizability tradeoffs in neural network training are driven by the interplay between training methods and network architecture. Employing optimization to train neural networks, which are typically over-parameterized to ensure universality, results in overfit networks that generalize poorly. In contrast, for statistical models with few parameters, optimization generically selects parameters that produce generalizable predictions. The effect of parametrization on the effectiveness of traditional training approaches suggests that systematically reducing neural networks' parameter spaces can improve model generalizability. Although we have primarily described sufficient training as an alternative to optimization-based training, it can also be seen as a model reduction technique. Past work
\cite{transtrumMBAM2016,mattinglyMaximizingInfo2018}
has shown that powerful model reduction techniques can be constructed by
leveraging the spectrum of the model's Fisher information metric on parameter
space. This same Fisher information metric spectrum underlies the power of the
simmering algorithm we introduced here.
As discussed in Methods, simmering traverses the loss landscape along
sloppy directions, collecting a minimally-biased ensemble of models that can
then be aggregated to average away the effect of sloppy directions and in turn
the effect of over-parameterization. The degree of reduction is modulated by the
temperature parameter $T=1/\beta$, whose effect on the Fisher information metric
spectrum is described in Methods. Therefore, one can start with an arbitrarily
over-parameterized network and aggregate the ensemble collected during sufficient
training to produce sufficiently simple models that capture the phenomenon of
interest without overspecification.

\clearpage
\printbibliography[title={References}, resetnumbers=false, segment=1]
\end{refsegment}

\section*{Methods}
\begin{refsegment}

\subsection*{Information Geometric Framing \label{sec:info_geo_framing}}
A neural network's training loss landscape is specified by its parameters $\vec{x}$, training dataset $\mathcal{D}$ and
loss function evaluated over the training data $L(\vec{x},\mathcal{D})$ where we will drop the vector notation on $x$
hereafter. The Pareto-Laplace transform of $L(x,\mathcal{D})$ is
\begin{equation}
     Z(\beta, \mathcal{D}) = \int d^N x e^{-\beta L(x,\mathcal{D})}, \label{eqn:partition_fn_x}
\end{equation}
where $\beta=1/T$ is the inverse temperature and $Z$ has the form of a general
partition function we are familiar with from statistical physics.
Now, for the given dataset, we define a set of collective variables $\vec{\theta}(\vec{x},\mathcal{D})$ (for which
we will also drop the vector notation) that are a function of the neural
network parameters and the training data. We can rewrite Eq.\ \ref{eqn:partition_fn_x} in terms of the
collective coordinates $\theta$ by first inserting the identity operator
\begin{equation}
    Z(\beta, \mathcal{D}) = \int d^n\theta
     \int d^N x e^{-\beta L(x,\mathcal{D})}
    \delta^n(\theta(x, \mathcal{D})-\theta),
    \label{eqn:partition_doubleint}
\end{equation}
and then rewriting Eq.\ \ref{eqn:partition_doubleint} only in terms of $\theta$
\begin{equation}
    Z(\beta, \mathcal{D}) = \int d^n\theta
     e^{-\beta F(\theta,\mathcal{D})}, \label{eqn:partition_fn}
\end{equation}
where
\begin{equation}
     e^{-\beta F(\theta,\mathcal{D})}=\int d^N x e^{-\beta L(x,\mathcal{D})}\delta^n(\theta(x, \mathcal{D})-\theta).
 \label{eq:LFEdef}
\end{equation}
The effective free energy, $F(\theta,\mathcal{D})$, specifies the probability density $p(\theta|\mathcal{D})$ of sampling a particular value of $\theta$. Up to an overall constant,
\begin{equation}
     p(\theta|\mathcal{D}) \propto e^{-\beta F(\theta,\mathcal{D})}, \label{eqn:theta_prob}
\end{equation}
and this probability can be computed via the mean-value theorem
\begin{equation}
    e^{-\beta F(\theta,\mathcal{D})} = \left<e^{-\beta L(\theta,{\mathcal{D})}}\right>\Omega(\theta,\mathcal{D}),
    \label{eq:LFEdef2}
\end{equation}
where $\Omega(\theta,\mathcal{D})$ is the volume of the $n$-dimensional hypersurface of $x$
along which $\theta(x, \mathcal{D})$ is constant, $\left<\cdot\right>$ denotes the mean
on that surface (also called an ensemble average), and $L$ is evaluated over the training dataset $\mathcal{D}$. Typically we refer to
$\Omega(\theta,\mathcal{D})$ in terms of the entropy $S(\theta,\mathcal{D})$ through the relation
$S = \ln\Omega$. Additionally, since $\theta$ is a set of collective variables, we can assume that $e^{-\beta L(\theta,\mathcal{D})}$ is slowly varying over $\Omega(\theta,\mathcal{D})$ such that $\left<e^{-\beta L}\right>\approx e^{-\beta\left< L \right>}$. Therefore, taking $L(\theta,\mathcal{D})$ as the loss from the
ensemble average in Eq.\ \ref{eq:LFEdef2}, and replacing $T=1/\beta$, we get the
relation for the free energy $F(\theta,\mathcal{D})$
\begin{equation}
    F(\theta,\mathcal{D})=L(\theta,\mathcal{D})-TS(\theta,\mathcal{D}).
    \label{eqn:free_energy}
\end{equation}
The most probable $\theta$ values (and thereby, the most probable neural network
parameter combinations) are those that maximize $p(\theta|\mathcal{D})$ (Eq.
\ref{eqn:theta_prob}) and thus minimize $F(\theta,\mathcal{D})$ (Eq.\ \ref{eqn:free_energy}). Note that the free
energy-minimizing $\theta$ does not minimize $L(x,\mathcal{D})$ (nor even $L(\theta,\mathcal{D})$)
because at finite temperature an entropic force $-T\partial_\theta S(\theta,\mathcal{D})$
drives $\theta$ away from minimal loss. The entropic drive away from minimal
loss is generic because there are generally more ways for a system (e.g., a
neural network) to have non-minimal loss than minimal loss (neural networks that
do not exhibit this property could be trivially trained by randomly selecting
weights and biases). This entropic force is key to the functionality of
sufficient learning -- by incorporating finite-temperature dynamics into
parameter updates, entropic forces systematically drive the learning
trajectory away from loss-minimizing parameters.

Although simmering generates non loss-minimizing distributions automatically and
generically, one can also identify the collective coordinates $\theta$ for
computing $F(\theta, \mathcal{D})$ for a particular neural network via techniques such as
information geometry.\cite{quinnInPCA2019}

To do this analysis, consider the Fisher information metric (FIM), which is
defined as 
\begin{equation}
    g_{\mu\nu}(\beta,\mathcal{D}) =
    -\left<\frac{\partial^2 (\beta F(\theta,\mathcal{D}))}{\partial \theta_\mu \partial \theta_\nu} \right>
    \;.
    \label{eqn:fim1}
\end{equation} 
For the present purposes, Eq.\ \eqref{eqn:fim1} can be computed in physics terms as a
generalized susceptibility according to
\begin{equation}
    g_{\mu\nu}(\beta,\mathcal{D}) =
    \frac{1}{Z}
    \int d^n\theta
    e^{-\beta F(\theta,\mathcal{D})}
    \frac{\partial^2 (\beta F(\theta,\mathcal{D}))}{\partial \theta_\mu \partial \theta_\nu} \; ,
    \label{eqn:fim_susceptibility}
\end{equation}
where $Z$ is defined in Eq.\ \eqref{eqn:partition_fn}.
The FIM has been interpreted in the work of Ref.\ \cite{quinnInPCA2019} as
an object that can be used to identify key ``order parameters'' that describe
data--model relationships. Ref.\ \cite{quinnInPCA2019} showed that the
spectrum of $g_{\mu \nu}$ yielded ``sloppy'' modes that describe parameters that
are loosely
constrained by data, as well as ``stiff'' modes that are highly constrained.
This mode classification provides a useful lens for interpreting the following
analysis of the effect of temperature on the probability distribution of
$F(\theta,\mathcal{D})$ (and subsequently, the neural network parameters sampled) in the
context of the Pareto-Laplace filter (Eq.
\ref{eqn:partition_fn_x}--\ref{eqn:free_energy}).  

Consider the parameter space near a free energy minimum such that
$\partial_{\theta} F=0$. For simplicity, take the minimum at $\theta=0$ (which
we can do without loss of generality by making a coordinate transformation). Taylor-expanding the FIM (Eq.\ \ref{eqn:fim_susceptibility}) near the minimum yields
\begin{equation}
    g_{\mu\nu}(\beta,\mathcal{D})
    \approx
    \frac{1}{Z}
    e^{-\beta F(0,\mathcal{D})}
    \int d^n\theta
    e^{-\frac{1}{2}\beta \theta_\alpha \theta_\gamma
    \left.\frac{\partial^2 F(\theta,\mathcal{D})}{\partial \theta_\alpha \partial \theta_\gamma}\right|_{\theta=0}}
    \frac{\partial^2 (\beta F(\theta,\mathcal{D}))}{\partial \theta_\mu \partial \theta_\nu} \; , \label{eqn:fim_expand}
\end{equation}
and
\begin{equation}
    Z
    \approx
    e^{-\beta F(0,\mathcal{D})}
    \int d^n\theta
    e^{-\frac{1}{2}\beta \theta_\alpha \theta_\gamma
    \left.\frac{\partial^2 F(\theta,\mathcal{D})}{\partial \theta_\alpha \partial \theta_\gamma}\right|_{\theta=0}}
    \; .
    \label{eqn:Z_expand}
\end{equation}
For sufficiently large $\beta$ (or low $T=1/\beta$), these expressions can be
integrated using a saddle point approximation, which yields
\begin{equation}
    g_{\mu\nu}(\beta,\mathcal{D})
    \approx
    \beta
    \left.\frac{\partial^2 F(\theta,\mathcal{D})}{\partial \theta_\mu \partial \theta_\nu}\right|_{\theta=0}
    \; .
    \label{eqn:fim_saddlepoint_approx}
\end{equation}
The FIM describes the relative ``proximity'' of $\theta$ to the free energy minimum $\theta=0$ in parameter space as a function of
$\beta=1/T$. The squared displacement $ds^2$ between $\theta$ and $\theta=0$ is
\begin{equation}
    ds^2 \propto
    \frac{1}{T} \sum_j \lambda_j d\lambda_j^2
    \; , \label{eqn:fe_eigenvalues}
\end{equation}
where $\lambda_j$ are the eigenvalues of the Hessian of $F(\theta,\mathcal{D})$ that describe the curvature of $F$ along different modes in
parameter space. Distances (Eq.\ \ref{eqn:fe_eigenvalues})
in a finite-temperature system diverge for small $T$,
except along directions corresponding to ``sloppy modes'' for which the
corresponding eigenvalues $\lambda_j\to 0$. Simmering
exploits this effect of temperature on distances in parameter space to
systematically sample families of loss-equivalent networks located along these
sloppy modes.

The Pareto-Laplace filter provides one of many possible mechanisms for generating ensembles of models at non-minimal loss. The Pareto-Laplace route we describe has the
advantage that, from an information theory point of view, it makes a minimal
assumption about the deviation between the empirical error-minimizing representation and the ground truth.\cite{ParetoLaplace}
There may be cases in which other information about the ground truth is available, in which case an
extension of the current approach via physics-informed neural networks \cite{PINNs}
is likely to provide minimally-biased sufficient-training methods.

This information geometric framing presented here also motivates the effectiveness of retrofitting at reducing overfitting in a neural network. The effect of $T$ on the Hessian of $F(\theta,\mathcal{D})$ (shown in Eq.\ \ref{eqn:fe_eigenvalues}) is analogous to the effect of the regularization strength parameter on the eigenvalues of the Hessian of the training loss in $\text{L}_2$-norm regularization.\cite{hastieStatisticalLearning2009} Via its temperature schedule, retrofitting rescales distances in parameter space in the same manner as $\text{L}_2$-norm regularization, which has already been shown to reduce overfitting.\cite{hastieStatisticalLearning2009}

\subsubsection*{Prediction uncertainty quantification with simmering}

In a generic supervised learning task, we aim to learn the ground truth based on an training dataset $\mathcal{D}=\{z_i,t_i\}^M_{i=1}$ comprised of a set of measured inputs $\{z_1,z_2,...,z_M\}$ and targets $\{t_1,t_2,...,t_M\}$. We use a neural network to model our estimation of ground truth $y(z^\prime;\vec{\theta})$ dependent on an unseen input $z^\prime$ and collective coordinates $\vec{\theta}(\vec{x},\mathcal{D})$ (where we will drop the vector notation as in the previous section). The probability density of predicting a target value $t^\prime$ based on an unseen input $z^\prime$ is 
\begin{equation}
    p(t^\prime|z^\prime, \mathcal{D})=\int d^n \theta \ p(t^\prime|z^\prime, \theta)p(\theta|\mathcal{D}). \label{eqn:predictive_dist}
\end{equation}
According to Bayes' rule, the posterior distribution $p(\theta|\mathcal{D})$ in Eq.\ \ref{eqn:predictive_dist} is
\begin{equation}
    p(\theta|\mathcal{D}) \propto p(\mathcal{D}|\theta)p(\theta), \label{eqn:bayes_rule}
\end{equation}
where $p(\theta|D)$ is the canonical distribution of $\theta$ at $\beta=1/T=1$.\cite{nealInference1993} Generalizing the posterior distribution to a general $T$, we recover Eq.\ \ref{eqn:theta_prob}-\ref{eq:LFEdef2}, with $F(\theta,\mathcal{D})$ defined in Eq.\ \ref{eqn:free_energy}. Comparison between Eq.\ \ref{eqn:theta_prob} and Eq.\ \ref{eqn:bayes_rule} identifies $S(\theta, \mathcal{D})$ as the regularizer on the collective coordinates $\theta$, and the temperature $T$ as the regularization strength. 

As in Section \ref{sec:info_geo_framing}, we consider the region of the parameter space near a free energy minimum, and for convenience this minimum occurs at $\theta=0$. We can expand $F(\theta,\mathcal{D})$ near this minimum as
\begin{equation}
    F(\theta, \mathcal{D}) \approx F(0, \mathcal{D}) +\frac{1}{2\beta}\left.g_{\alpha\gamma}(\beta, \mathcal{D})\right|_{\theta=0}{\theta_\alpha\theta_\gamma}, \label{eqn:free_energy_approx}
\end{equation}
using Eq.\ \ref{eqn:fim_saddlepoint_approx} to relate $\frac{\partial^2 F(\theta, \mathcal{D})}{\partial \theta_\alpha \partial \theta_\gamma}$ to $g_{\alpha\gamma}(\beta, \mathcal{D})$ for large $\beta$ (small $T$). Using Eq.\ \ref{eqn:theta_prob} and \ref{eqn:free_energy_approx}, we can rewrite $p(t^\prime|z^\prime,\mathcal{D})$ as
\begin{equation}
    p(t^\prime|z^\prime,\mathcal{D}) \propto \int d^n \theta \ p(t^\prime|z^\prime,\theta)e^{-\frac{1}{2} \left.g_{\alpha\gamma}(\beta, \mathcal{D})\right|_{\theta=0}{\theta_\alpha\theta_\gamma}}. \label{eqn:general_predictive_dist}
\end{equation}
Eq.\ \ref{eqn:general_predictive_dist} shows that for a generic training dataset $\mathcal{D}$, simmering reduces distances in parameter space, allowing for sampling of near-optimal $\theta$ values that can elucidate features of $p(t^\prime|z^\prime,\theta)$. 

In the case where the training data targets have additive Gaussian noise $\epsilon \sim \mathcal{N}(0,\sigma^2_t)$, and we choose mean-squared error (MSE) as the loss function, $p(t^\prime|z^\prime,\theta)$ becomes
\begin{equation}
    p(t^\prime|z^\prime,\theta)= {\bigg( \frac{1}{2\pi\sigma^2_t}\bigg)}^{\frac{1}{2}} e^{-\frac{1}{2\sigma^2_t}{(y(z^\prime;\theta)-t^\prime)}^2}.\label{eqn:target_uncert}
\end{equation}
Then, we can expand $y(z^\prime;\theta)$ around the free energy-minimizing $\theta=0$ as
\begin{equation}
    y(z^\prime;\theta)\approx y(z^\prime;0)+\left.\frac{\partial y(z^\prime;\theta)}{\partial \theta_\alpha}\right|_{\theta=0}\theta_{\alpha}.\label{eqn:pred_taylor_expand}
\end{equation}
Using Eq.\ \ref{eqn:target_uncert} and the Taylor-expansion of $y(z^\prime;\theta)$ in Eq.\ \ref{eqn:pred_taylor_expand}, we can evaluate Eq.\ \ref{eqn:general_predictive_dist} to be
\begin{equation}
    p(t^\prime|z^\prime,\mathcal{D})\propto e^{-\frac{{(t^\prime-y(z^\prime;0) )}^2}{2\sigma^2}}
\end{equation}
where $\sigma^2=\sigma^2_t+ \left.g_{\alpha\gamma}(\beta, \mathcal{D})^{-1}\right|_{\theta=0}\left.\frac{\partial y(z^\prime;\theta)}{\partial \theta_\alpha}\right|_{\theta=0}\left.\frac{\partial y(z^\prime;\theta)}{\partial \theta_\gamma}\right|_{\theta=0}$. In this case, the prediction uncertainty distribution is similar to that of Bayesian inference with noise injection,\cite{matsuokaNoiseInjection1992,wrightBayesianNN1999} but generalized to any inverse temperature $\beta$. The variance $\sigma^2$ of the prediction uncertainty distribution has two terms: the variance introduced by uncertainty in data target measurements, $\sigma^2_t$, and a variance-like quantity defined by the sensitivity of
the free energy and the model predictions to changes in $\theta$. The overall effect of temperature on the second variance term is to reduce the FIM, $g_{\alpha \gamma}$, thereby increasing the second variance term contribution, but the exact way in which changing $g_{\alpha\gamma}(\beta,\mathcal{D})$ affects $\partial_\theta y$ is problem-dependent. 

\subsection*{Parameters as a system of particles}
The Pareto-Laplace transform of the neural network loss in Eq.\ \eqref{eq:PLxfm}
yields a generating function for neural network parameters. This generating
function has the form of a partition function in statistical mechanics, which creates the possibility to generate networks by adapting molecular simulation
methods.

To employ molecular dynamics techniques in a neural network problem, we treat
the neural network parameters as a system of one-dimensional particles in an interaction
potential. The value of each weight and bias defines the position of each
particle in the physical system, and the loss function acts as the system
potential. The negative gradient of the training loss function acts as a force on the
system of particles that pushes them towards a minimum of the loss function.
If we also define a conjugate momentum for each particle, we can integrate the equations of motion for the physical system and iteratively generate sets of weights and
biases. In the absence of other forces, this integration would yield a set of
weights and biases of increasing accuracy on the training data, and as such is
analogous to the result of any traditional gradient-descent training algorithm.

Modelling the neural network parameters as a physical system in this way allows
us to apply ensemble sampling methods from statistical physics to collect an
ensemble of models.

\subsection*{Nos\'e-Hoover Chain Thermostat}

A thermostat, in a molecular dynamics context, is an algorithm that controls the
temperature of a physical system.\cite{frenkelsmit} In simmering, we use a
Nos\'e-Hoover chain (NHC) thermostat, but other thermostats can also be employed
to achieve constant-temperature conditions. The use of a thermostat allows us to
sample from the canonical ensemble of neural network parameters, and thus
produce ensembles of models at different temperature scales. The NHC thermostat
samples from the canonical ensemble by introducing an interaction between the
neural network parameters and a chain of massive virtual
particles.\cite{frenkelsmit} The first particle in the chain exchanges energy
with the system of neural network parameters, and the rest of the chain only interacts with their neighbouring chain particles. Each virtual particle has a position and
a conjugate momentum that are computed iteratively along with those of the system of real
particles.

Given a set of $N$ neural network parameters, we define a set of positions
$x=\{x_i\}$, associated momenta $\{p_i\}$ and masses $\{
m_i\}$. Using this set of quantities, we can model a system of
one-dimensional particles in a potential defined by a loss function
$L(x, \mathcal{D})$ that depends on the neural network parameter positions
(weights and biases) $x$ and the training dataset $\mathcal{D}$. This physical system is also interacting with
an NHC of length $N_c$, where each constituent particle also has its own mass $Q_k$,
position $s_k$ and momentum $p_k$. Henceforth, the neural network parameters
will be referred to as the ``real'' particles, to contrast with the virtual
particles of the NHC.

The Hamiltonian of this coupled system at temperature $T_{\text{target}}$ is  \cite{GlennMartyna1996,Tuckerman1992} 
\begin{align}
    \mathcal{H} &= \mathcal{H}_{system} + \mathcal{H}_{NHC} \nonumber \\
    &= \sum^{N}_{i=1}\frac{1}{2} \frac{{p_i}^2}{m_i} + L(x, \mathcal{D}) + \sum^{N_c}_{k=1} \frac{1}{2}\frac{{p_k}^2}{Q_k} +NT_{\text{target}}s_1 + \sum^{N_c}_{k=2}T_{\text{target}}s_k. \label{eqn:hamiltonian}
\end{align}

For simplicity of notation, we will henceforth set $m_i=1,Q_k=1 \ \forall \ i,k$
and describe the integration process in terms of the positions and velocities,
rather than the positions and momenta. The equations of motion are derived from
the Hamiltonian, and are given by
\begin{align}
    \dot{x}_i(t) &= v_i(t) \label{eqn:eom_1} \\
    \dot{v}_i(t) &= a_i(t) - v_{s_k}(t)v_i(t) \label{eqn:eom_2} \\
    \dot{v}_{s_1}(t) &= a_{s_1}(t) - v_{s_2}(t)v_{s_1}(t) \label{eqn:eom_3} \\
    \dot{v}_{s_k}(t) &= a_{s_k}(t) - v_{s_{k+1}}(t)v_{s_k}(t), \label{eqn:eom_4}
\end{align}
where the $i$ subscript denotes real particle quantities, and the $s_k$
subscript denotes a quantity affiliated with the $k^{th}$ virtual particle. The
accelerations of the particles are given by
\begin{align}
    a_i(t) &= -\frac{1}{m_i} \nabla L(x_i,\dots ,x_N) \label{eqn:real_acc}\\
    a_{s_1}(t) &= \frac{1}{Q_1}\big(\sum_i m_i v^2_i - (N - N_e - N_{ine})T_{target}\big) \label{eqn:virt_as1}\\
    a_{s_k}(t) &= \frac{1}{Q_k}\big(Q_{k-1} m_i v^2_{k-1} - T_{target}\big). \label{eqn:virt_ask}
\end{align}
The acceleration of the real particles in Eq.\ \ref{eqn:real_acc} is proportional
to the negative gradient of the loss with respect to the weights and biases in
the neural network. Given
Equations \ref{eqn:eom_1}--\ref{eqn:virt_ask}, the trajectories of the real and virtual
particles, and thus the evolution of neural network weights and
biases over training time, can be determined numerically.

Further details on implementation and model architecture are given in SI.

\section*{Data Availability}
No data were generated in the course of this investigation.

\section*{Code Availability}
Code for implementing simmering is available open source at Ref.\
\supercite{SimmeringCode}.

\printbibliography[category={methods},title={Methods References}, resetnumbers=false]
\end{refsegment}

\section*{Acknowledgements}
We thank C.X.\ Du, A.A.\ Klishin, M.\ Spellings, and J.\ Wammes for discussions
and P.\ Chitnelawong, K.\ Huneau, and M.\ Sheahan for collaboration at an early
stage of this work. We acknowledge the support of the Natural Sciences and
Engineering Research Council of Canada (NSERC) grants RGPIN-2019-05655 and
DGECR-2019-00469. IB acknowledges the support of the NSERC USRA program, a
Canada Graduate Scholarship, and the Vector Institute.

\section*{Author Contributions}
IB, HA, and GvA designed research. IB contributed original code and data. IB,
HA, and GvA analyzed results. IB, HA, and GvA wrote the paper. GvA initiated and
supervised research.

\section*{Competing Interests}
The authors have no competing interests to declare.

\clearpage

\section*{Supplementary Information}

\subsection*{Numerical implementation}

To discretize the equations of motion Eqs.\ \ref{eqn:eom_1}--\ref{eqn:virt_ask}, we use a Verlet
integrator for position and velocity for both the real and virtual particles. Verlet integration is used here because it preserves Hamilton's equations.$^\text{39,40}$ The resulting discretized equations of motion are
\begin{align}
    x_i(t+\Delta t/2) &= x_i(t) + \frac{\Delta t}{2}v_i(t) \label{eqn:verlet_x1} \\
    s_{2k}(t+\Delta t/2) &=  s_{2k}(t) + \frac{\Delta t}{2}v_{s_{2k}}(t) \label{eqn:verlet_seven}\\
    v_{s_{2k-1}}(t + \Delta t/2) &= v_{s_{2k-1}}(t)e^{\frac{-\Delta t}{2}v_{s_{2k}}}+ \frac{\Delta t}{2}a_{s_{2k-1}}(t)e^{\frac{-\Delta t}{4}v_{s_{2k}}}, \label{eqn:verlet_oddv1} \\
v_i(t+\Delta t) &= v_i(t)e^{-\Delta t v_{s_1}(t+\Delta t/2)} +\Delta t \ a_i(t+\Delta t/2)e^{-\frac{1}{2}\Delta t v_{s_1}(t+\Delta t/2)}\label{eqn:verlet_v} \\
s_{2k-1}(t+\Delta t) &= s_{2k-1}(t)+\Delta tv_{s_{2k-1}}(t+\Delta t/2) \label{eqn:verlet_sodd}\\
v_{s_{2k}}(t+\Delta t) &= v_{s_{2k}}(t)e^{-\Delta t v_{s_{2k+1}}(t+\Delta t/2)}+\Delta ta_{s_{2k}}(t+\Delta t/2)e^{-\frac{1}{2}\Delta t v_{s_{2k+1}}(t+\Delta t/2)}, \label{eqn:verlet_evenv} \\
x_i(t+\Delta t)&=x_i(t+\Delta t/2)+\frac{\Delta t}{2}v_i(t+\Delta t)\label{eqn:verlet_x2}\\
s_{2k}(t+\Delta t)&=s_{2k}(t+\Delta t/2)+\frac{\Delta t}{2}v_{s_{2k}}(t+\Delta t)\label{eqn:verlet_evens2}\\
v_{s_{2k-1}}(t+\Delta t) &= v_{s_{2k-1}}(t+\Delta t/2)e^{-\frac{1}{2}\Delta t v_{s_{2k}}(t+\Delta t)} +\frac{\Delta t}{2} a_{s_{2k-1}}(t+\Delta t)e^{-\frac{1}{4}\Delta t v_{s_{2k}}(t+\Delta t)}. \label{eqn:verlet_evenv2}
\end{align} 

Given this numerical integration scheme, an ensemble of neural networks can be generated by selecting an appropriate learning rate $\Delta t$ and iteratively computing $x(t + \Delta t)$.

\subsection*{Model System Architecture}

The examples shown in the main text demonstrate the performance of simmering on a variety of neural network problems. Different architectures, activation functions, loss functions and initializations are used to highlight the breadth of contexts in which sufficient training can be applied.

We implemented simmering using the TensorFlow
library in Python. For each test case, the built-in
training step in TensorFlow is replaced with the Verlet integration scheme described in
Equations \ref{eqn:verlet_x1}--\ref{eqn:verlet_evenv2} to iteratively produce
sets of weights and biases. The acceleration $a_i(t+\Delta t/2)$ in Equation
\ref{eqn:verlet_v} for each iteration is supplied by the automatic differentiation
of the loss with respect to the weights and biases in TensorFlow. In all cases, full-batch gradient descent is used, so the batch size is taken
to be the size of the entire dataset.

We have published an open-source version of the simmering code, which is available at Ref.\ 38. The
published code allows for retrofitting and ab initio sufficient training to be
conducted on the noisy sine dataset shown in Fig.\ \ref{fig:Retro} in the main text.

\subsubsection*{Retrofitting Overfit Networks}
To retrofit a neural network, we use the final set of (overfit) weights produced
by a traditional optimization algorithm and a first order approximation of the neural network
parameters' final velocities as the retrofitted neural network's initial
conditions. The final parameter velocities can be computed using the last and second-last iterations' weights,
\begin{equation}
    v_{i,t} \approx \frac{x_{i,t}-x_{i,t-1}}{\Delta t} \label{eqn:velocity_approx},
\end{equation}
where $\Delta t$ is the learning rate.
We use both the final weights and velocities to ensure that the
neural network is initialized in the exact location in phase space where the
optimizer stopped. We then define a temperature schedule for the thermostat,
starting at $T=0$ as the network was not coupled to a thermostat during optimization. The
published code has the option to implement a temperature schedule that increases
in a step-wise manner at equal intervals from $T=0$ to the user's target
temperature choice. Fig.\ \ref{fig:Retro}a shows an example of this
temperature schedule, and this type of step-wise temperature change was used for
all examples shown in panels Fig.\ \ref{fig:Retro}e-f. Once the target temperature is reached, we collect a finite-temperature ensemble to generate
ensemble predictions. For the classification examples shown in Fig.\ \ref{fig:Retro}e,
majority voting was used to aggregate the classifier predictions. For the
regression examples in Fig.\ \ref{fig:Retro}c,d,f, the ensemble samples were averaged to obtain the
ensemble prediction.

In each example shown in this work, the overfit networks were produced by training with the Adam optimizer, with its default
parameters in TensorFlow, and a constant learning rate of $\Delta t=0.002$.

Three different datasets were used to produce the classification results in Fig.\ \ref{fig:Retro}e: the MNIST handwritten
digits dataset \cite{deng2012_mnist}, the Higgs dataset \cite{whitesonHIGGS2014},
and the Iris dataset \cite{iris_53}.

The network architecture for learning the MNIST dataset was a LeNet-5
convolutional neural network \cite{lecunGradient1998}, modified to have ReLU
activations rather than sigmoid activations for all layers. Categorical
cross-entropy was used as the loss function. 10,000 images were used for
training, and 1000 for testing, both of which were selected randomly without
overlap from the dataset. The network weights were initialized using the Glorot
normal initializer in TensorFlow, and the network was trained using the Adam optimizer for 300 epochs. During simmering, the thermostat temperature was
increased from $T_{\text{initial}}=0$ to $T_{\text{target}}=0.001$ in steps of
$\Delta T = 0.001$ every 200 iterations, with a learning rate of $\Delta t =
0.002$. Simmering was conducted for 6,000 iterations, and the last 4,000
iterations' neural network parameters were used for ensemble predictions. 

For the Higgs dataset example, the network architecture consisted of 4 512-unit hidden layers, and a linear output layer. The hidden layers were given an exponential linear unit (ELU) activation. The loss, which
was chosen to be binary cross-entropy, was computed from logits to account for
the linear output. 10,000 samples were used for training, and 1000 samples were
used for testing, selected randomly with no overlap from the Higgs dataset. The
network weights were initialized using the TensorFlow Glorot uniform
initializer. The network was trained using the Adam optimizer for 1000 epochs.
During simmering, the thermostat temperature was increased from
$T_{\text{initial}}=0$ to $T_{\text{target}}=0.0001$ in steps of $\Delta T =
0.0001$ every 500 iterations, with a learning rate of $\Delta t = 0.002$. Simmering was carried out for 15,000 iterations, and the models resulting from the last 12,000 iterations of training were used
for producing ensemble predictions.

For the Iris dataset example, the network architecture consisted of 3 hyperbolic
tangent-activated hidden layers, and a linear output layer. The first hidden
layer had 100 units,and the subsequent two hidden layers had 50 units. The categorical
cross-entropy was used as the network loss. The Iris dataset consists of four
input features, but we used only the ``sepal width'' and the ``petal width'' features to classify
the flowers. 112 samples were used for training, and 38 were used for testing,
partitioned randomly without overlap. The input features were linearly rescaled
based on the training data features to map to the range $[-1,1]$. The network weights
were initialized using the TensorFlow Glorot normal initializer, and trained
using the Adam optimizer for 200 epochs with no batching. During simmering, the
thermostat temperature was increased from $T_{\text{initial}}=0$ to
$T_{\text{target}}=0.1$ in steps of $\Delta T = 0.01$ every 200 iterations, with
a learning rate of $\Delta t = 0.002$. The simulation was carried out for
10,000 iterations, and 10\% of the last 7000 iterations were used for ensemble
predictions.

Two datasets were used to produce the regression retrofitting examples shown in Fig.\ \ref{fig:Retro}f: a noisy
sine curve and the Auto-MPG dataset \cite{quinlanAutoMPG1993}. For both regression
datasets, ``min-max'' scaling was used during training, which performs a linear
transformation on the input data such that all features map to the range $[-1,
1]$. The parameters of the linear transformation are set before training based
on the training data feature scales. The inverse of this linear scaling is
applied to the model output before it is compared with the training data targets. 

The noisy sine curve data was generated by adding normally distributed noise to
a sine curve in the following manner:
\begin{equation}
    y = \sin(2\pi x) + 0.1\ \mathcal{N}(\mu=0,\sigma=1). \label{eqn:noisy_sine}
\end{equation}
The inputs used were a set of 101 equally spaced points over the interval
$[-1,1]$. The dataset was split into a training set of 65 points and a test set
of 36 points, partitioned randomly with no overlap. The neural network
used  had two 20-unit hidden layers, and 1
output node. The two hidden layers had a hyperbolic tangent activation, and the
output layer had a linear activation. The loss function used was
sum-squared error (SSE). The network weights were initialized using the built-in
TensorFlow Glorot normal initializer. The Adam optimizer was used to train the
network weights for 2000 epochs. During simmering,
the thermostat temperature was increased from $T_{\text{initial}}=0$ to
$T_{\text{target}}=0.05$ in steps of $\Delta T = 0.01$ every 1000 iterations,
with a learning rate of $\Delta t = 0.002$. The simulation was conducted for
10,000 iterations total, and the final 3,000 iterations contributed to
the ensemble prediction shown in Fig.\ \ref{fig:Retro}d.

The Auto-MPG dataset was used for both single input (S) and multivariate (M)
regression. In the single input case, only the ``horsepower'' feature was
used to predict the target (miles per gallon), but in the multivariate case, all six
features in the dataset to predict the same target (miles per gallon). For the single
input case, the neural network architecture consisted of 2 ReLU-activated
64-unit hidden layers, and a linear output layer. The train/test split for the
Auto-MPG dataset was 80/20 (313 train samples, 79 test samples) partitioned
randomly with no overlap. The loss function used was SSE. The network weights
were initialized using the TensorFlow Glorot normal initializer, and trained for
3500 epochs with the Adam optimizer. During simmering, the thermostat temperature was increased
from $T_{\text{initial}}=0$ to $T_{\text{target}}=0.4$ in steps of $\Delta T =
0.1$ every 200 iterations, with a learning rate of $\Delta t = 0.001$. The
simulation was conducted for 12,000 iterations total, and the final 6,000
iterations were selected for use in the ensemble prediction.

For the multivariate Auto-MPG training problem, the neural network architecture consisted of 2
hyperbolic tangent-activated 64-unit hidden layers, and a linear output layer.
The train/test split for the Auto-MPG dataset was 315 train/77 test,
partitioned randomly with no overlap. The loss function used was mean-squared
error (MSE). The network weights were initialized using the TensorFlow Glorot
normal initializer, and trained for 1500 epochs with the Adam optimizer. During simmering, the thermostat temperature was increased from
$T_{\text{initial}}=0$ to $T_{\text{target}}=0.5$ in steps of $\Delta T = 0.1$
every 200 iterations, with a learning rate of $
\Delta t = 0.002$. The
simulation was conducted for 10,000 iterations total, and the final 6,000
iterations were selected for use in the ensemble prediction.

\subsubsection*{Ab Initio Sufficient Training}
For ab initio sufficient training, simmering was employed at the outset and a constant temperature was maintained for the entire duration of the training process.

To generate the classification example, the Iris dataset was used. The choice of dataset
partitioning, network architecture and loss function were the same as in the
retrofitting example. The weights were initialized using a Glorot normal
initialization, and simmering was employed from the outset at a thermostat
temperature of $T=0.002$ for 25,000 iterations with a learning rate of $\Delta
t=0.001$. Over the last 10,000 iterations, 2,000 model samples are randomly selected
for the ensemble prediction. Simmering was conducted on models
with 36 different random seeds (resulting in distinct instances of the Glorot normal initializations) while keeping the
data train/test partition fixed, and the ensemble majority prediction was computed based
on the total of votes across from all 36 replications' sampled models. In
Figure \ref{fig:AbInitioSimmering}a, the result of the ensemble majority vote decision boundary is shown.
The background colour in Figure \ref{fig:AbInitioSimmering}a is a weighted average of the three class
colours, and reflects what proportion of the ensemble of models voted for each
Iris species. 

For the regression example, the Auto-MPG dataset was used. As in the single
variable retrofitting example, the ``horsepower'' feature was used to
predict the target (miles per gallon). The dataset was partitioned such that there
were 300 training samples and 92 test samples, categorized randomly with no
overlap. In this case, the network architecture consisted of 1 10-unit
hyperbolic tangent-activated hidden layer, and a linear output. MSE was used as
the loss function. The weights were initialized using a modified Glorot normal
initialization. The overall distribution of weights for each layer has the same mean and width
as the corresponding Glorot normal distribution, but the range
$[-2\sigma,2\sigma]$ is split into $n$ equal segments (where $n$ is the number
of input nodes to that layer), and each weight's value is generated from a
normal distribution centred on the midpoint of one of the segments. Simmering
was employed from the outset at a thermostat temperature of $T=1$ for 40,000
iterations with a learning rate of $\Delta t=0.002$. In Figure \ref{fig:AbInitioSimmering}b, the ensemble
predictions are plotted every 40 iterations past 1000 iterations, but all 39,000
iterations are used to produce the ensemble prediction and the uncertainty
distributions shown in Figures \ref{fig:AbInitioSimmering}e and \ref{fig:AbInitioSimmering}f.

\printbibliography[category={si},title={Supplementary Information References}, resetnumbers=false]
\end{document}